\documentclass[10pt,twocolumn,letterpaper]{article}

\usepackage[pagenumbers]{cvpr} 

\usepackage{graphicx}
\usepackage{amsmath}
\usepackage{amssymb}
\usepackage{booktabs}
\usepackage{bm}

\usepackage{amsmath,amsfonts,bm}

\def\eqref#1{equation~\ref{#1}}

\def\1{\bm{1}}

\DeclareMathAlphabet{\mathsfit}{\encodingdefault}{\sfdefault}{m}{sl}
\SetMathAlphabet{\mathsfit}{bold}{\encodingdefault}{\sfdefault}{bx}{n}

\newcommand{\R}{\mathbb{R}}

\usepackage[pagebackref,breaklinks,colorlinks]{hyperref}

\usepackage[capitalize]{cleveref}
\crefname{section}{Sec.}{Secs.}
\Crefname{section}{Section}{Sections}
\Crefname{table}{Table}{Tables}
\crefname{table}{Tab.}{Tabs.}

\begin{document}

\title{SSGVS: Semantic  Scene Graph-to-Video Synthesis}

\author{Yuren Cong\textsuperscript{1}, Jinhui Yi\textsuperscript{2}, Bodo Rosenhahn\textsuperscript{1}, Michael Ying Yang\textsuperscript{3}\\
\textsuperscript{1}TNT, Leibniz University Hannover,
\textsuperscript{2}University of Bonn,
\textsuperscript{3}SUG, University of Twente\\
}

\maketitle

\begin{abstract}
As a natural extension of the image synthesis task, video synthesis has attracted a lot of interest recently.
Many image synthesis works utilize class labels or text as guidance. 
However, neither labels nor text can provide explicit temporal guidance, such as when an action starts or ends. 
To overcome this limitation, we introduce semantic video scene graphs as input for video synthesis, as they represent the spatial and temporal relationships between objects in the scene.
Since video scene graphs are usually temporally discrete annotations, we propose a video scene graph (VSG) encoder that not only encodes the existing video scene graphs but also predicts the graph representations for unlabeled frames.
The VSG encoder is pre-trained with different contrastive multi-modal losses.
A semantic scene graph-to-video synthesis framework (SSGVS), based on the pre-trained VSG encoder, VQ-VAE, and auto-regressive Transformer, is proposed to synthesize a 
video given an initial scene image and a non-fixed number of semantic scene graphs. 
We evaluate SSGVS and other state-of-the-art video synthesis models on the Action Genome dataset and demonstrate the positive significance of video scene graphs in video synthesis.
The source code will be released.
\end{abstract}

\section{Introduction}
\label{sec:intro}

\begin{figure*}[t!]
\begin{center}
\includegraphics[width=0.8\linewidth]{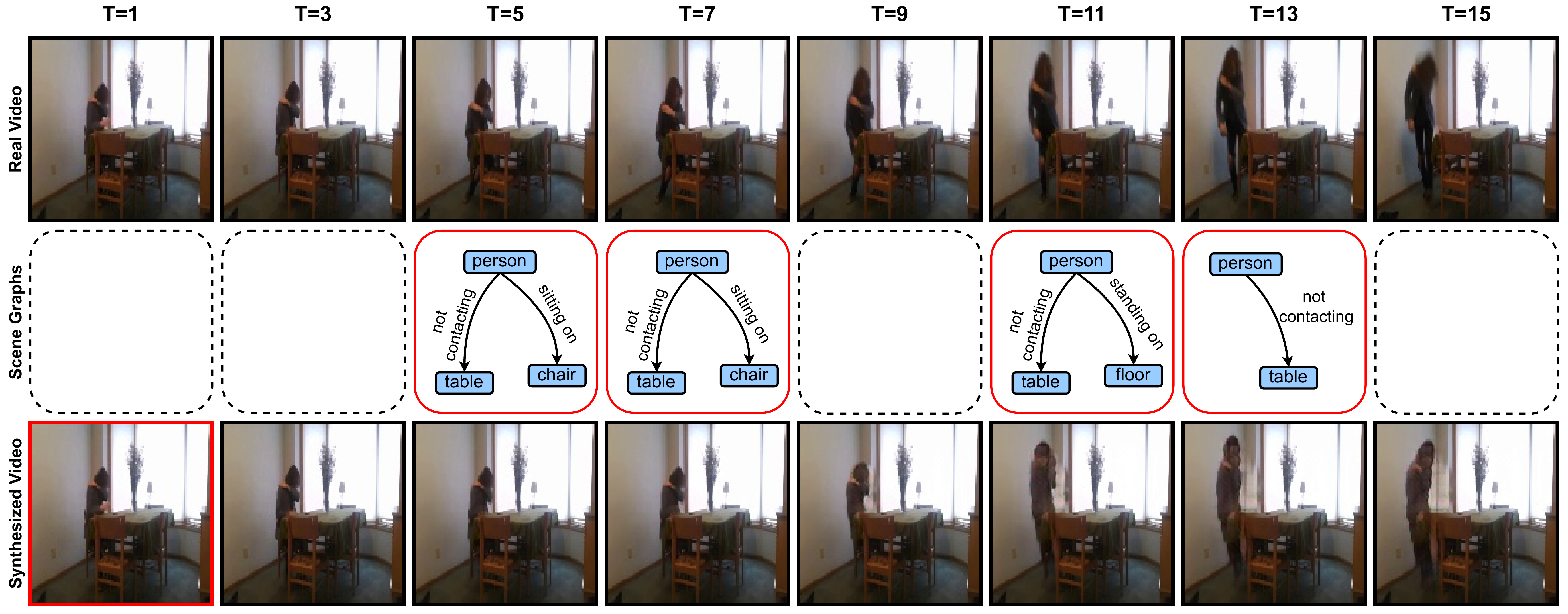}
\end{center}
\vspace{-3mm}
\caption{Given a starting frame and a non-fixed number of semantic scene graphs (with red bounding boxes), SSGVS can synthesize a fix length complex video with $128\times128$ resolution, not just simple repetitive motions. The number and temporal position of the input video scene graphs are freely controllable. For clarity, only 8 of 16 frames are shown.}
\label{fig:teaser}
\vspace{-5mm}
\end{figure*}

With the tremendous breakthroughs in image synthesis \cite{karras2020training, saharia2022photorealistic, yu2022scaling} in recent years, more and more researchers are focusing on the natural extension yet more challenging task of video synthesis.
In contrast to image synthesis, video synthesis requires that each generated frame has spatial fidelity and also that these frames conform to temporal continuity. 
To generate a semantically meaningful video, the guidance is essential. 
Text-guided image generation models have achieved convincing performance, and there has been recent works \cite{li2018video, ho2022video, hong2022cogvideo} of video synthesis using text as condition.
However, text has the drawback of not providing explicit guidance for time dependencies.
For example, when we want to generate a 16-frame video from the text condition ``a girl standing up after sitting on a chair", the synthesis model needs not only to generate frames with visual quality and temporal consistency, but also to infer which of the 16 frames correspond to the ``sitting" state and which to the ``standing up" action.
This highly increases the complexity of video synthesis task.
AG2Vid \cite{bar2020compositional} introduced continuous action graphs for video synthesis, which specify semantic actions and spatial layouts of objects.
Although action graphs contain explicit spatial-temporal information, it is demanding to artificially define the layout variations of all objects as input when generating real scene videos.

Is there a balanced guiding condition that contains semantic information and clearly represents temporal dependencies?
Based on this observation, we make use of semantic video scene graphs for the task of video synthesis.
An image scene graph is a structural representation that generalizes the objects of interest in an image as nodes and their relationships as edges. 
It has been utilized as conditional information for image generation \cite{johnson2018image}.
Semantic video scene graphs can be viewed as several static scene graphs on an temporal axis, which only describe the semantic content of the frames in the video.
The ideal video scene graphs are continuous. 
In other words, each frame in the video is annotated with a corresponding scene graph.
This data structure containing spatial and temporal information is very promising for video synthesis.
However, giving continuous video scene graphs during inference is challenging compared to giving other temporal conditions such as trajectories \cite{le2021ccvs}, action labels \cite{menapace2021playable}, or overall scene conditions such as text, because scene graphs are relatively difficult to create.

In this paper, we propose a novel semantic scene graph-to-video synthesis framework (SSGVS) that can synthesize a fixed-length video with an initial scene image and discrete semantic video scene graphs, as shown in \cref{fig:teaser}. 
To address the discontinuity of the input video scene graphs and to learn graph representations to better guide video synthesis, we propose a video scene graph (VSG) encoder pre-trained in a contrastive multi-modal framework.
The VSG encoder can not only provide the graph representations for the existing video scene graphs but also predicts the representations of the frames that are missing a video scene graph.
Therefore, our approach does not impose strict restrictions on the number and temporal location of the given video scene graphs, nor does it require additional spatial information, which makes SSGVS more feasible in practice.
For the generative model, we utilize a popular combination of VQ-VAE and auto-regressive Transformer \cite{yan2021videogpt,le2021ccvs, ge2022long} since this likelihood-based model fits our purpose and is easy to optimize. 
The graph representations can be inserted into the sequence of the discrete latents provided by the encoder of VQ-VAE.
These latents and graph representations are then modeled by a GPT-like Transformer using an auto-regressive prior.
The latents generated from the auto-regressive prior are converted to video frames by the decoder of VQ-VAE. 
Our main contributions are listed as follows:
\begin{itemize}
\item We propose a novel semantic scene graph-to-video synthesis framework (SSGVS), which can synthesize a fixed-length video with an initial scene image and discrete  semantic video scene graphs. 

\item A contrastive multi-modal learning framework is proposed to pre-train a Transformer-based video scene graph (VSG) encoder which can provide high-quality video scene graph representations to condition semantic video synthesis. 


\item We split a sub-dataset from the  Action Genome dataset \cite{ji2020action} and conduct experiments to demonstrate the benefits of using semantic video scene graphs as condition for video synthesis. Compared to other state-of-the-art works, SSGVS can generate better quality semantic videos.
\end{itemize}


\begin{figure*}[http!]
\begin{center}
\includegraphics[width=0.8\linewidth]{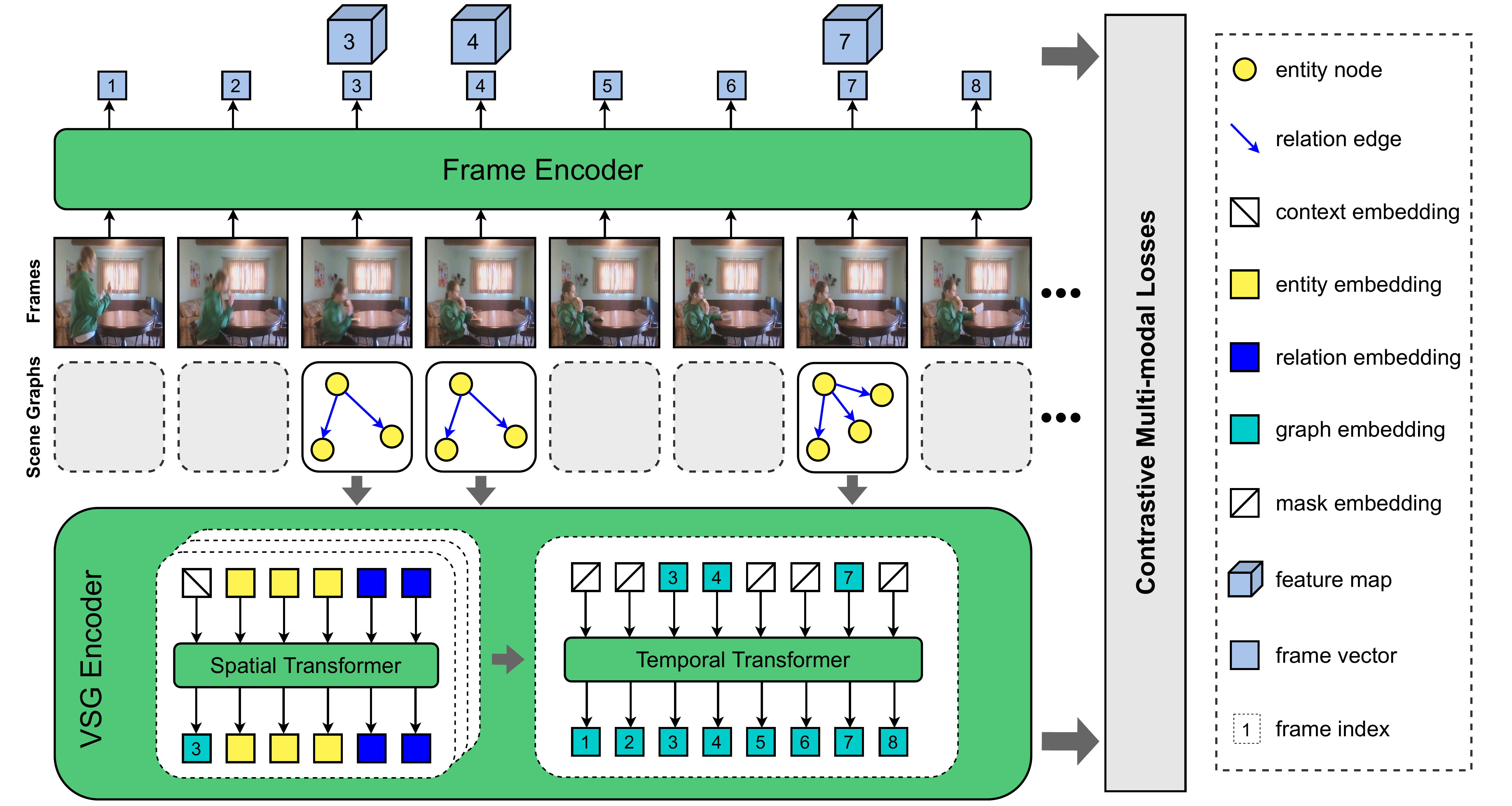}
\end{center}
\vspace{-3mm}
\caption{Our video scene graph representation learning framework, which connects semantic video scene graphs to video frames. The VSG encoder converts the given video scene graphs into single-vector representations and reasons about the representations of the ungiven scene graph, which are further used for video synthesis. }
\vspace{-3mm}
\label{fig:vsg_pretraining}
\end{figure*}

\section{Related Work}
\label{Related_work}

\noindent
\textbf{Scene graph.}
Scene graphs have first been proposed by \cite{johnson2015image} for image retrieval and have received a lot of attention in the field of scene understanding. 
The graphical representation whose nodes indicates objects and edges indicates the semantic relationships between objects can clearly describe the scene. 
There are many impressive 
works \cite{zellers2018neural, tang2019learning, lin2020gps, chiou2021recovering, li2021bipartite, dhingra2021bgt, lyu2022fine,  liu2022constrained, liu2021fully, li2022sgtr, teng2022structured, cong2022reltr} that have achieved incredible results of image scene graph generation on the datasets \cite{lu2016visual, krishna2017visual, kuznetsova2020open}.
Recently, the dataset Action Genome \cite{ji2020action} is proposed, which extends image scene graphs to video scene graphs by adding the temporal dimension.
Some works \cite{cong2021spatial,li2022dynamic,xu2022meta} captures spatial and temporal dependencies and generates video scene graphs. 
Since scene graphs not only contain the objects present in the scene but also demonstrates the interactions between the objects, they are exploited in image retrieval \cite{johnson2015image}, visual question answering \cite{damodaran2021understanding} and image synthesis \cite{johnson2018image, yang2022scene,herzig2020learning}.
Video scene graphs have also been used for video question answering \cite{cherian20222} and video captioning \cite{cao2020using}.
In this paper, we use video scene graphs for video synthesis, which are purely semantic and do not include spatial information of entities.

\noindent
\textbf{Video synthesis.}
A video can be regarded as a high-dimensional image with an additional temporal dimension.
Therefore, the methods for image synthesis can be extended to video synthesis as well, although longer training time and higher memory consumption are required. 
Some GAN-based methods adapt the adversarial framework to generate videos with 3D convolutions \cite{acharya2018towards, wang2020g3an, menapace2021playable} or recurrent neural networks \cite{tulyakov2018mocogan, clark2019adversarial, saito2020train}.  However, as the length of the generated videos becomes longer, the quality of the videos synthesized by these models decreases significantly. 
Auto-regressive models \cite{kalchbrenner2017video, weissenborn2019scaling} have become popular in this field, even though their inference speed is slow. Many of these methods \cite{wu2021n, yan2021videogpt, le2021ccvs, ge2022long, hong2022cogvideo} introduce VQ-VAE \cite{van2017neural} and Transformers \cite{vaswani2017attention} to improve performance.
Furthermore, some works \cite{ho2022video, yang2022diffusion} propose video diffusion models which is a natural extension of the image architecture. 
There are many sub-tasks in the field of video synthesis. 
In unconditional synthesis, a video is produced without any prior information. For conditional synthesis, a popular task is to leak a few video frames to the model, which predicts future videos.
Some aforementioned works support the generation of videos conditional on class labels or semantic labels \cite{yu2022modular}, while text-to-video models \cite{balaji2019conditional,hong2022cogvideo,ho2022video} are rapidly developing.
To further control the video content, trajectory of a robotic arm \cite{le2021ccvs} and camera motions \cite{ren2022look} are exploited.
AG2Vid \cite{bar2020compositional} introduced continuous action graphs for video synthesis.
However, the scene layout and spatial variations are necessary input, which limits the application.
In this paper, we introduce semantic video scene graphs as the guidance.
We use a scene image and purely semantic video scene graphs as the input of an auto-regressive model to synthesize a video. 
Although video scene graphs are usually temporally discrete, the proposed video scene graph encoder pre-trained with different contrastive losses can predict continuous video scene graph embeddings for guiding video synthesis.
Different from previous works, SSGVS allows precise control of the moment when the episodes occur, while the input is relatively easy to achieve.
\section{Video scene graph representation learning}
\label{vsgrl}
Many text-to-image generation works \cite{ramesh2022hierarchical, saharia2022photorealistic} using CLIP text encoder \cite{radford2021learning} have achieved outstanding results.
It demonstrates that pre-trained condition encoders provide high-quality representations that facilitate the generation task.
With the same motivation, we propose a video scene graph-video contrastive pre-training framework including a video scene graph (VSG) encoder and a frame encoder (see \cref{fig:vsg_pretraining}). The pre-trained VSG encoder converts the given video scene graphs into single-vector representations and reasons about the representations of the ungiven scene graph. The auxiliary frame encoder, that is discarded during synthesis, provides the frame representations and feature maps. We establish a mapping of the semantic graph representation space to the visual latent space by using the graph-frame representation similarity while the feature maps provide fine-grained information at the node and edge level.

\subsection{Video scene graph encoder}
\label{vsgencoder}
Our video scene graph encoder consists of a spatial Transformer and a temporal Transformer.
The spatial transformer captures the context of the input scene graphs and the fine-grained representations of their nodes and edges, while the temporal Transformer infers the graph representations at all times based on the graph context provided by the spatial Transformer.

\vspace{2mm}
\noindent
\textbf{Spatial Transformer.} 
Given a scene graph $G=(\bm{N}, \bm{E})$, the nodes $\bm{N}$ and edges $\bm{E}$ in the graph are viewed as tokens with semantic information.
We decompose the structure of the graph into a sequence $\bm{S}$ consisting of node tokens and edge tokens. 
In order to obtain the graph context through self attention mechanism, we introduce a special \texttt{[context]} token and place it always first in the sequence $\bm{S}$. 
Each token corresponds to a learned embedding $\bm{e} \in \displaystyle \R^{d}$ that is randomly initialized before being trained from scratch.
In order for the structural properties of the graph to be preserved in Transformer which is permutation invariant, we construct the context encoding $\bm{E}(c) \in  \displaystyle \R^{d}$ and node encodings $\bm{E}(\bm{N})$ with learned embeddings. 
For the edge $e_{ij}$ from the node $n_i$ to the node $n_j$, the edge encoding is calculated as $\bm{E}(e_{ij})= \bm{E}(n_i)-\bm{E}(n_j)$. 
These encodings are element-wise added to the corresponding embeddings.
For a graph with $N_n$ nodes and $N_e$ edges, the length of the input sequence is $(1+N_n+N_e)$.
%
We adopt a GPT-like multi-layer Transformer in this paper. Each transformer layer consists of a classical multi-head attention module, a feed-forward network, and normalization layers. 
We use the original full attention mechanism but not sparse attention in Transformers, which is defined as:
\vspace{-1mm}
\begin{equation}
\centering
\begin{aligned}
  Attention(\bm{Q},\bm{K},\bm{V}) = Softmax\left(\frac{\bm{Q}\bm{K}^T}{\sqrt{D_k}}\right) \bm{V},
\end{aligned}
\vspace{-1mm}
\end{equation}
where $\bm{Q}$, $\bm{K}$, and $\bm{V}$ are the linear transformations of the queries, keys and values, respectively. 
The feed-forward network is a two-layer perceptron, while layer normalization is used in the Transformers for normalization.
The graph context from the last layer of the spatial Transformer is forwarded to the temporal Transformer.
The node and edge representations are used to learn fine-grained visual information from the feature maps provided by the frame encoder with the fine-grained graphical contrastive loss introduced in \cref{cmml}.

\vspace{2mm}
\noindent
\textbf{Temporal Transformer.}
Although the temporal Transformer has the identical architecture as the spatial Transformer, it plays a completely different role, inferring graph representations for unavailable scene graphs. 
A special \texttt{[mask]} token is introduced whose learned embedding has the same dimension as the graph context.
We construct a sequence of the mask tokens with the same length $T$ as the video and and replace the mask embeddings in the position of the given graphs with their graph context provided by the spatial Transformer. 
Temporal encodings $E_t \in \displaystyle \R^{T\times d} $ are customized to inject temporal location information into the sequence. These encodings are also learned embeddings.
This sequence is used as the input to the temporal Transformer that reasons about the representations of all graphs based on temporal dependencies. 
Our motivation is similar to the masked language model\cite{devlin2018bert} that masks some words in the sentence and reconstructs them using the context. 
The difference is that these graph representations $\bm{g} \in \displaystyle \R^{d}$ are learned with the contrastive learning using the frame vectors provided by the frame encoder.

\subsection{Frame encoder}
\label{frame_encoder}
To relate the graph representations to the visual appearance, we utilize a frame encoder to encode the video frames and extract the feature maps.
It is built upon a pre-trained convolutional neural network \cite{szegedy2016rethinking}. 
A $1\times1$ convolution layer is exploited to reduce the dimension of the feature maps extracted by the pre-trained CNN.
The frame vectors $\bm{v}\in \displaystyle \R^d$ are derived from the feature maps $\in \displaystyle \R^{h\times w \times d}$ through a 2D global average pooling layer and a linear transformation.
The frame vectors of all frames are involved in the loss calculation, while only the feature maps of the frames with graph annotations are activated for fine-grained learning.

\begin{figure*}[http!]
\begin{center}
\includegraphics[width=0.8\linewidth]{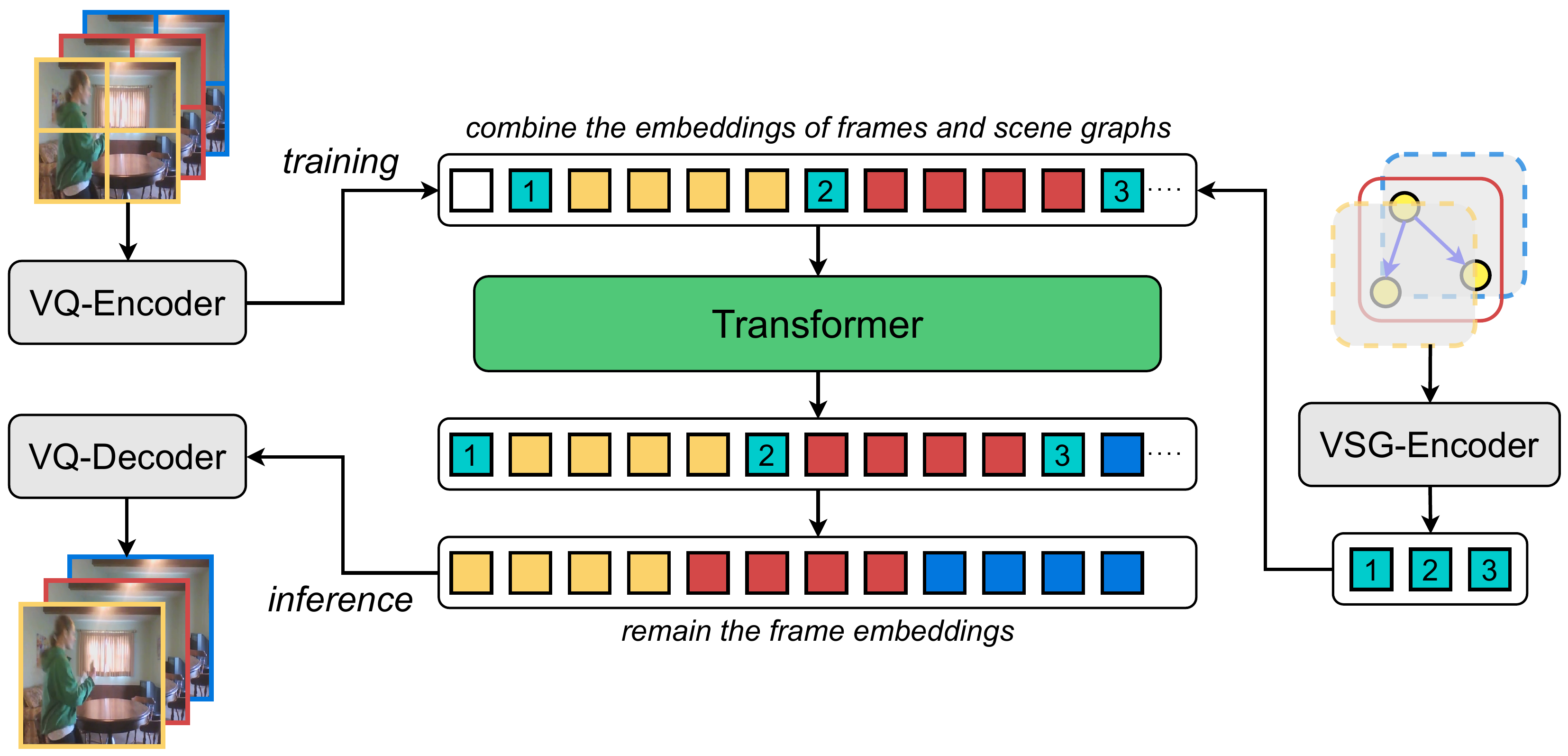}
\end{center}    
\vspace{-3mm}
\caption{Overview of our semantic scene graph-to-video synthesis framework SSGVS. During training, Transformer learns the prior distribution of the sequence consisting of the latent embeddings provided by VQ-VAE and the scene graph embeddings provided by the VSG encoder. In inference, the latent embeddings of the starting frame and graph embeddings are given. The latent embeddings of future frames are sampled from the prior and decoded into the video by VQ-VAE.}
\vspace{-5mm}
\label{vsg-vsf}
\end{figure*}

\subsection{Contrastive multi-modal losses}
\label{cmml}
We introduce three contrastive loss functions for graph representation learning, namely graphical intra-video contrastive loss, graphical inter-video contrastive loss, and graphical fine-grained contrastive loss. The total loss function is the equal sum of these three loss functions.

\vspace{2mm}
\noindent
\textbf{Graphical intra-video contrastive loss.} 
Temporal dependencies and continuity are present in both visual appearance and semantic representations. 
Therefore, we propose to learn scene graph representations $\bm{g}$ from visual representations $\bm{v}$ through temporal dependencies. We compute the graphical intra-video contrastive loss of a video with $T$ frames as:
\vspace{-3mm}
\begin{equation}
\begin{split}
L_{intra}=-\sum_{i=1}^T (\log\frac{\exp(\textrm{sim}(\bm{g}_i,\bm{f}_i))}{\sum_{j=1}^T \exp(\textrm{sim}(\bm{g}_j,\bm{f}_i))} 
\\+ \log\frac{\exp(\textrm{sim}(\bm{g}_i,\bm{f}_i))}{\sum_{j=1}^T \exp(\textrm{sim}(\bm{g}_i,\bm{f}_j))}),
\end{split}
\label{eq:graphical_intra_loss}
\vspace{-3mm}
\end{equation}
where $\textrm{sim}(\bm{g}_i,\bm{f}_j)$ indicates the cosine similarity between the graph representation of the $i$-th frame and the frame vector of the $j$-th frame. 
The first term 
denotes that the frame vector should have a higher similarity to the graph representation of the current frame compared to the graph representation of other frames, while the second term is symmetrical for the graph representation.

\vspace{2mm}
\noindent
\textbf{Graphical inter-video contrastive loss.} 
Because a scene graph is the semantic description of a scene, videos with different visual appearances correspond to different scene graphs.
The objective here is to learn graph representations by comparing scenes from different videos. 
Given a batch of $B$ videos, the graphical inter-video contrastive loss of the $t$-th frame can be formulated as:
\vspace{-1mm}
\begin{equation}
\centering
\begin{split}
L_{inter}^t=-\sum_{i=1}^B (\log\frac{\exp(\textrm{sim}(\bm{g}_i,\bm{f}_i))}{\sum_{j=1}^B \exp(\textrm{sim}(\bm{g}_j,\bm{f}_i))} 
\\+ \log\frac{\exp(\textrm{sim}(\bm{g}_i,\bm{f}_i))}{\sum_{j=1}^B \exp(\textrm{sim}(\bm{g}_i,\bm{f}_j))}),
\end{split}
\label{eq:graphical_inter_loss}
\vspace{-3mm}
\end{equation}
where the frame index $t$ on the right side of the equation is omitted for brevity.
The graph representation $\bm{g}_i$ and frame vector $\bm{f}_j$ are inferred from the the $t$-th frames of the $i$-th video and the $j$-th video in the batch.
The complete $L_{inter}$ are the sum of $L_{inter}^t$ of $T$ frames.

\vspace{2mm}
\noindent
\textbf{Graphical fine-grained contrastive loss.}
Inspired by \cite{xu2018attngan}, we propose a fine-grained loss function that supervises the consistency between the frame and the semantic nodes as well as edges of the scene graph.
The graph representations are improved at a fine-grained level by associating the node and edge representations with the sub-regions in the video frame.
We view both the node and edge representations of the scene graph as semantic embeddings $\bm{e}$. For a scene graph with $N_n$ nodes and $N_e$ edges, there are totally $N_g=N_n+N_e$ semantic embeddings. We compute the visual context $\bm{c}_i$ of the $i$-th node or edge in the graph as:
\begin{equation}
\centering
\begin{split}
\bm{c}_i = \sum_{j=1}^{hw} \frac{\exp(\textrm{sim}_{dot}(\bm{e}_i, \bm{r}_j))}{\sum_{k=1}^{hw} \exp(\textrm{sim}_{dot}(\bm{e}_i, \bm{r}_k))} \bm{r}_j, 
\\ \textrm{where } \textrm{sim}_{dot}(\bm{e}_i, \bm{r}_j)=\frac{\exp(\bm{e}_i^\textrm{T} \bm{r}_j)}{\sum_{g=1}^{N_g}\exp(\bm{e}_g^\textrm{T} \bm{r}_j)}.
\end{split}
\label{eq:dot_similarity}
\end{equation}
$\bm{r}_j \in \displaystyle \R^{d}$ is the vector of the $j$-th sub-region in the feature map with $h\times w$ shape provided by the frame encoder, whereas $\textrm{sim}_{dot}(\bm{e}_i, \bm{r}_j)$ denotes the normalized dot-product similarity between the $i$-th semantic embedding of the scene graph and the visual vector of the $j$-th sub-region. 
The matching score $\textrm{S}(G, F)$ between the scene scene $G$ and the frame $F$ can be formulated as $\textrm{S}(G, F)=\log(\sum_{i=1}^{N_g}\exp(\textrm{sim}(\bm{e}_i, \bm{c}_i))$.
When there are $P$ pairs of video frames and scene graphs in the batch of $B$ videos, the graphical fine-grained contrastive loss is defined as:
\begin{equation}
\centering
\begin{split}
L_{finegrain}=-\sum_{i=1}^P (\log\frac{\exp(\textrm{S}(G_i,F_i))}{\sum_{j=1}^P \exp(\textrm{S}(G_j, F_i))} 
\\ + \log\frac{\exp(\textrm{S}(G_i,F_i))}{\sum_{j=1}^P \exp(\textrm{S}(G_i,F_j))}).
\end{split}
\label{eq:graphical_inter_loss}
\end{equation}

\section{Semantic scene graph-to-video synthesis}
\label{VSGVS}
In this section, we introduce a semantic scene graph-to-video synthesis framework (SSGVS) consisting of a video scene graph (VSG) encoder, a VQ-VAE, and an auto-regressive Transformer. 
The overview of SSGVS is illustrated in \cref{vsg-vsf}.

\noindent
\textbf{VQ-VAE.}
We adopt the VQ-VAE proposed by \cite{le2021ccvs} and pre-train it using videos in the training set. 
Given a frame, the CNN-based encoder produces the output with the down-sampled spatial resolution. The discrete latent variables $\bm{z}$ for the output are quantified by the nearest neighbor look-up using the shared embedding space.
The latent embeddings are forwarded to the decoder to reconstruct the input frame. 
We freeze the pre-trained VQ-VAE in other stages.

\vspace{2mm}
\noindent
\textbf{Auto-regressive Transformer.}
We utilize an auto-regressive Transformer with full attention mechanism to learn a prior over the VQ-VAE latent embeddings $\bm{z}$ and video scene graph representations $\bm{g}$. 
The basic architecture of the auto-regressive Transformer is almost identical to the Transformers in \cref{vsgencoder}, but with more layers.
During the training stage, the latent embeddings of the video frames are flattened frame by frame into a 1D sequence in row-major order.
The learned temporal encodings are added to the latent embeddings.
Then we insert the graph representations provided by the VSG encoder into the sequence. 
Note that each frame has a corresponding scene graph representation.
For frames where no scene graph is given, the VSG encoder is responsible for inferring the representations.
To prevent information leakage, an empty embedding $\bm{z}_0$ is given at the beginning of the sequence by convention. 
The Transformer learns to model the prior distribution of $\bm{z}$ by minimizing the negative log-likelihood for latent codes as:
\begin{equation}
\centering
L_{latent}=\mathbb{E}_{\bm{z}\sim \textrm{p}(\bm{z}_{data})}(-\log \textrm{p}(\bm{z})).
\end{equation}
Although scene graph embeddings are known in the inference stage, we also optimize the mean square error loss between the input and output graph embeddings.

During the inference, the starting frame and video scene graphs are given. 
The discrete latent embeddings of the starting frame are computed by the VQ-encoder, while the VSG encoder infers the scene graph representations. 
The initial sequence is constructed with the empty embedding, these latent embeddings, and graph representations.
The latent embeddings of the future frames are randomly sampled from the categorical distribution learned by the auto-regressive Transformer.
The VQ-decoder synthesizes the video with the known and predicted latent embeddings.
There are three approaches to inserting graph representations: 1) Insert the graph representation in front of the latent embeddings of the corresponding frame as shown in \cref{vsg-vsf}; 2) insert the graph representation after the latent embeddings of the corresponding frame; 3) insert all graph representations in front of the latent embeddings of all frames. We use the first approach because it has the best performance.

\section{Experiments}
\label{experiments}

\noindent
\textbf{Dataset and evaluation metrics.}
We split a sub-dataset from the video scene graph dataset Action Genome \cite{ji2020action}, which is built upon the In-Home dataset Charades \cite{sigurdsson2016hollywood}.
There are $70k$ training videos and $7k$ test videos with 36 object categories and 17 relationship categories. 
Each video with a resolution of 128$\times$128 contains 16 frames and a variable number of video scene graphs.
The maximum number of nodes in scene graphs is set to 5.
We sampled every 5 frames from the original videos to capture large motion. 

To evaluate the model performance, we adopt the Fr\'{e}chet Video Distance (FVD) proposed by \cite{unterthiner2018towards} that estimates the distribution distance between real and synthesized videos in the feature space.
Following \cite{le2021ccvs}, we calculate the
mean and standard deviation of FVD in 5 evaluations.
Furthermore, we evaluate the structural similarity index measure (SSIM) \cite{wang2004image} which evaluates the similarity between the original and synthetic frames.

\vspace{2mm}
\noindent
\textbf{Implementation details.}
%
%
In the video scene graph encoder, both the spatial Transformer and temporal Transformer have 3 Transformer layers. 
We employ 4 attention heads for each attention module, while the dimension $d$ of the input queries, keys, and values is set to 256.
We train the video scene graph encoder and frame encoder using ADAM optimizer \cite{kingma2014adam} with a learning rate of $1\times10^{-4}$ and a batch size of 12 images.
The training takes about 20 hours on 2 RTX 2080 TI GPUs.

We train the VQ-VAE using ADAM optimizer \cite{kingma2014adam} with a learning rate of $2\times10^{-4}$ and a batch size of 32 videos on 8 RTX 3090 TI GPUs for about 48 hours.
The auto-regressive Transformer consists of 24 Transformer layers with a head number of 16. 
Due to the complexity of the auto-regression task, the embedding dimension of the attention module is set to 1024.
Therefore, a linear transformation is utilized to project the dimension of video scene graph representations from 256 to 1024, while 1024 latent embeddings with dimension $1024d$ are learned during the training.
We train the auto-regressive Transformer using ADAM optimizer \cite{kingma2014adam} with a learning rate of $1\times10^{-5}$ and a batch size of 64 videos on 8 RTX 3090 TI GPUs for about 48 hours. 
Furthermore, the video scene graph encoder is frozen during the training of the auto-regressive Transformer.
Refer to supplementary material for more details about the hyperparamters and model architecture.
\begin{table}[t!]
\caption{Comparison with state-of-the-art video synthesis methods on the split Action Genome. Given the starting frame, a 16-frame video with $128\times128$ resolution is synthesized. For our method SSGVS, video scene graphs are provided as additional conditions. With the help of the VSG encoder, SSGVS achieves the best performance in all metrics.$\checkmark$ denotes video scene graphs are provided. }
\label{tab:quantative}
\begin{center}
\vspace{-3mm}
\begin{tabular}{lccc}
\hline
\multicolumn{1}{c}{\bf Method}  & \multicolumn{1}{c}{\bf Graph} &\multicolumn{1}{c}{\bf FVD ($\downarrow$)} &\multicolumn{1}{c}{\bf SSIM ($\uparrow$)} \\
\hline
MoCoGAN  \cite{tulyakov2018mocogan} &-  &$911.2\pm18.6$ &0.459 \\
LVT \cite{rakhimov2020latent} &-   &$572.7\pm25.4$ & 0.493 \\
VideoGPT \cite{yan2021videogpt} &-  &$888.6\pm19.7$ & 0.472 \\
CCVS \cite{le2021ccvs}   &- &$426.7\pm21.4$ &0.516\\
SSGVS (Ours)  &\checkmark &\bm{$382.2\pm15.2$} & \textbf{0.565} \\
\hline
\end{tabular}
\end{center}
\vspace{-7mm}
\end{table}

\begin{figure*}[http]
\begin{center}
\includegraphics[width=0.95\linewidth]{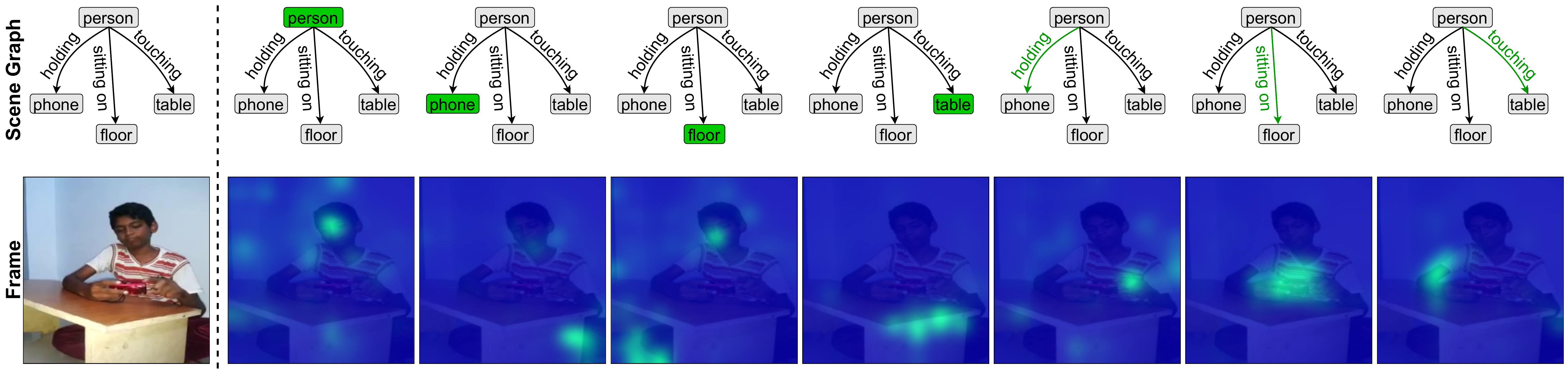}
\end{center}
\vspace{-3mm}
\caption{Attention maps of the node and edge representations to sub-regions in the frame. The node or edge in the scene graph corresponding to the attention map in the second row is highlighted in green. \texttt{phone} cannot match the correct region due to the small bounding box and low resolution.}
\label{fig:finegrain_attention}
\vspace{-3mm}
\end{figure*}

\subsection{Comparison with state-of-the-art methods}
To the best of our knowledge, there is no previous work that synthesizes videos from semantic video scene graphs.
To verify whether semantic scene graphs have a positive effect on video synthesis, we evaluated the performance of some advanced models on our dataset that can synthesize future frames given a starting frame.
Please note that we have selected only those works whose official code is published, and the amount of computation required to reproduce them is within our capabilities.
\cref{tab:quantative} demonstrates that our model SSGVS outperforms other state-of-the-art methods that only use the first frame to synthesize a video.
With the help of video scene graphs, FVD between the real videos and videos synthesized by SSGVS is $382.2$, which is $44.5$ lower than CCVS and $190.5$ lower than LVT.
For SSIM, SSGVS also has the best performance.
In addition, when we train the models on the sub-dataset split from Action Genome, VideoGPT does not perform  well.
We speculate that this is because Action Genome dataset is more complex than the widely-used datasets for video synthesis \cite{soomro2012ucf101, ebert2017self}.
The scenes are diverse, the camera pose is not fixed, and the motions in the videos are large.

\subsection{Graph representation learning analysis} 
In contrast to the previous generative methods, we introduce video scene graphs as a condition to guide video synthesis.
In order to clarify how graph representation learning contributes to the synthesis performance, we first ablate different contrastive multi-modal losses and present the results in \cref{tab:ablate_loss}.
The first row indicates that the video scene graph encoder is integrated into SSGVS without pre-training and optimized with the auto-regressive transformer.
The FVD score increases significantly from $382.2$ to $457.1$.
In this case, the quality of the generated videos is even worse than if only the first frame was given.
We conjecture that video scene graph representations cannot be learned without a reasonably designed loss function.
In particular, it is simultaneously optimized when training a complex generative model.
Although we pre-train the VSG encoder with only the graphical intra-video contrastive loss, FVD rapidly drops to 403.1.
Because the VSG encoder learns temporal dependencies, which are crucial for graph representation inference. 
The graphical inter-video contrastive loss also helps the VSG encoder pre-training, while graph representation quality can be further improved by using the graphical fine-grained contrastive loss. 
The FVD score decreases to 382.2, while SSIM increases to $0.565$.
\begin{table}[t]
\centering
\caption{We evaluate the synthesize performance by integrating the VSG encoders pre-trained with different contrastive loss functions in SSGVS. The first row indicates that the VSG encoder is not pre-trained, but optimized in the training of the auto-regressive Transformer. The second to fourth rows denote different losses are activated ($\checkmark$) in the pre-training.}
\begin{tabular}{ccccc}
\hline
$L_{intra}$ &$L_{inter}$ &$L_{finegrain}$ &\textbf{FVD ($\downarrow$)}  &\textbf{SSIM ($\uparrow$)} \\
\hline
- &- &- &457.1 $\pm$  10.6 & 0.509 \\
\checkmark&- &- &403.1 $\pm$  12.7  & 0.541\\
\checkmark &\checkmark &- &395.1 $\pm$ 16.1  & 0.551\\
\checkmark &\checkmark &\checkmark &\textbf{382.2 $\pm$ 15.2} & \textbf{0.565} \\
\hline
\label{tab:ablate_loss}
\end{tabular}
\vspace{-5mm}
\end{table}
\begin{table}[t]
\centering
\caption{Ablation study of graph representations. $\O$ indicates video scene graphs are not provided.  The second to fourth rows show the different representation insertion order used by SSGVS, the numbers correspond to the order numbers in \cref{VSGVS}.}
\begin{tabular}{ccc}
\hline
Order &\textbf{FVD ($\downarrow$)} &\textbf{SSIM ($\uparrow$)} \\
\hline
\O &426.7 ± 21.4 &0.516 \\
1  &\textbf{382.2 ± 15.2} & \textbf{0.565} \\
2  &391.8 ± 17.5 & 0.553 \\
3  &387.2 ± 13.1 & 0.559 \\
\hline
\label{tab:ablate_order}
\end{tabular}
\vspace{-5mm}
\end{table}



To demonstrate how the node and edge representations are associated with regions of interest using the graphical fine-grained contrastive loss, an instance of the attention weights in \cref{eq:dot_similarity} is visualized in \cref{fig:finegrain_attention}.
The first row shows the scene graph with different highlighted nodes and edges, while the second row shows the corresponding attention maps.
With our contrastive multi-modal learning, not only node representations can be localized to entities, but also semantic relationships can be projected to critical regions.
For example, \texttt{holding} and \texttt{touching} are usually closely associated with the hand or arm of \texttt{person}.

\begin{figure*}[http]
\begin{center}
\includegraphics[width=0.95\linewidth]{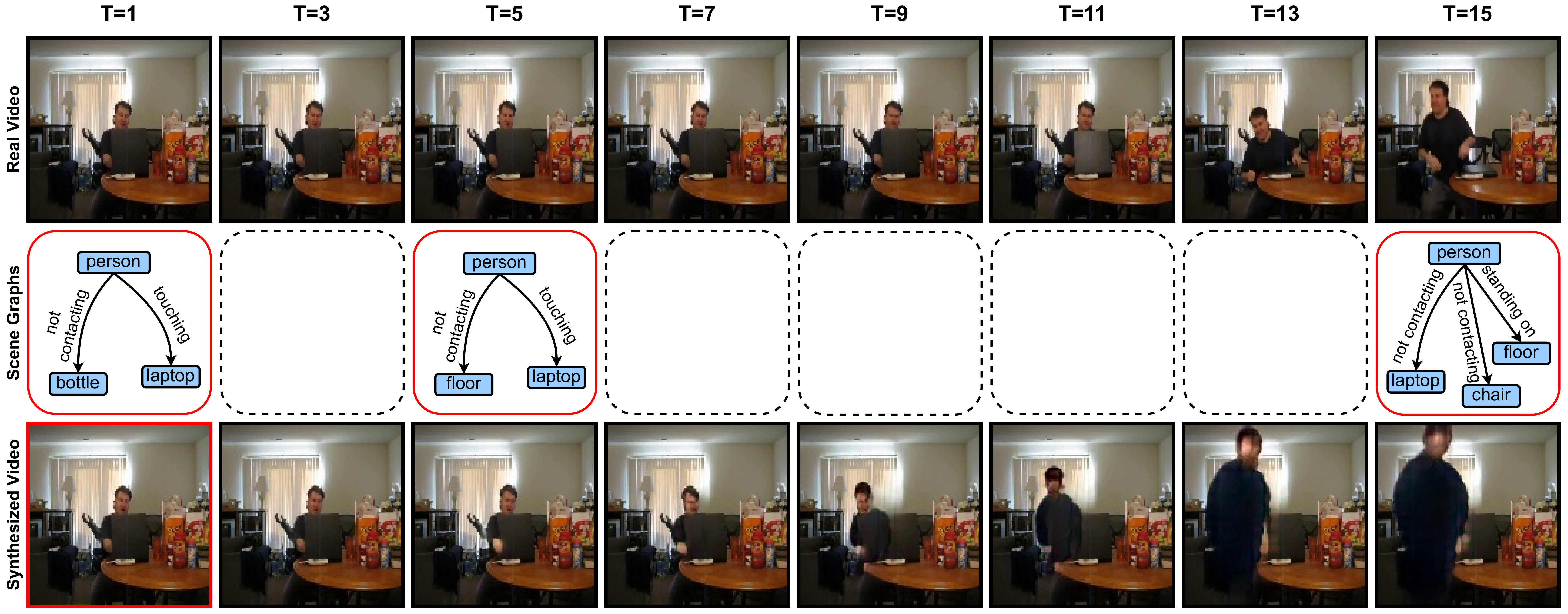}
\end{center}
\vspace{-3mm}
  \caption{Qualitative result of complex semantic video synthesis. Given the first frame and a several video scene graphs (with red bounding boxes), SSGVS can synthesize not only simple videos with small or repetitive movements, but also long-term videos with complex semantic content.}
\label{fig:qualitative}
 \vspace{-3mm}
\end{figure*}

\subsection{Ablation study} 

To verify the effect of video scene graphs on video synthesis, we ablate the video scene graph encoder during the inference. 
In addition, we evaluate the performance of SSGVS using different insertion graph representation orders introduced in \cref{VSGVS}:
1) Insert the graph representation in front of the latent embeddings of the corresponding frame; 2) insert the graph representation after the latent embeddings of the corresponding frame; 3) insert all graph representations in front of the latent embeddings of all frames.
The results are shown in \cref{tab:ablate_order}.
The first row indicates no scene graphs are given, while the auto-regressive Transformer predicts latent embeddings only based on the first frame.
Without video scene graphs, it is difficult for SSGVS to infer what will happen in the future and FVD decreases to 426.7.
Furthermore, the insertion order of graph representations also effects the synthesis performance, although it is not very critical.
SSGVS performs best when inserting the graph representation in front of the latent embeddings of the corresponding frame.

To explore the optimal model structure, we train the auto-regressive Transformer with different Transformer layers.
The results are shown in Tab. \ref{tab:ablate_layer}. 
We used the same settings for all experiments except for the number of layers $=32$. 
For the Transformer with 32 layers, the batch size is reduced to 24 videos due to GPU memory limitation.
In this case, the FVD score decreases to 410.4. 
There are two possibilities, either the model is too complex causing overfitting, or the optimization using a smaller batch size does not perform well.
Finally, we adopt the auto-regressive Transformer with 24 Transformer layers which has the best performance in practice.
%
\begin{table}[http!]
\caption{Ablation study of Transformer layers in the auto-regressive Transformer.}
\vspace{-3mm}
\begin{center}
\begin{tabular}{cc}
\hline
Layer number &\textbf{FVD ($\downarrow$)} \\
\hline
4 &476.0 ± 16.3 \\
8  & 422.9 ± 17.5 \\
16  &399.1 ± 13.7 \\
24  &\textbf{382.2 ± 15.2} \\
32  &410.4 ± 20.4 \\
\hline
\label{tab:ablate_layer}
\end{tabular}
\end{center}
\vspace{-7mm}
\end{table}

\subsection{Qualitative results}
\cref{fig:qualitative} shows the qualitative result for  semantic video synthesis from video scene graphs. 
The three rows from top to bottom are respectively the original video, video scene graphs, and the synthetic video.
As discussed in \cref{vsgrl}, the given scene graphs are discrete.
Given the first frame and several scene graphs (with red bounding boxes), 15 future frames are synthesized by SSGVS, while the even columns are omitted to save space.
SSGVS can synthesize semantically controllable videos, in this example the \texttt{person} is \texttt{sitting} and using a \texttt{laptop}, then \texttt{stands} up. 
Different from the original video, the \texttt{standing} action starts at $T=9$ instead of $T=15$.
Since the command \texttt{standing} is given in the scene graph at $T=15$, SSGVS infers that the complete standing will take more time based on the learned knowledge.
In the original video, the person is not yet fully standing up.
The difference could be eliminated by giving more video scene graphs as constraints.
However, some visual details, such as the face, are not well rendered in the last few generated frames, because generating long-term videos with large motions is very challenging.
For videos with small motion, SSGVS can render better details.
Due to space limitation, we present more qualitative results and discuss the limitations of SSGVS in the supplementary.

\section{Conclusion}
In this paper, we propose a semantic scene graph-to-video synthesis framework SSGVS which aims to synthesize complex semantic videos. Through contrastive multi-modal learning, our video scene graph encoder can infer continuous graph representations based on the given discrete scene graphs. Given the starting frame and graph representations as constraints, the latent embeddings of future frames are sampled from the distribution learned by the auto-regressive Transformer and converted to frames by the VQ-VAE. 
Our experiments demonstrate that video scene graphs have a positive effect on video synthesis.

\appendix

\section*{Appendix}

\section{Transformer architecture}
\label{appendix_transformer}
\begin{figure}[http!]
\begin{center}
\includegraphics[width=0.8\linewidth]{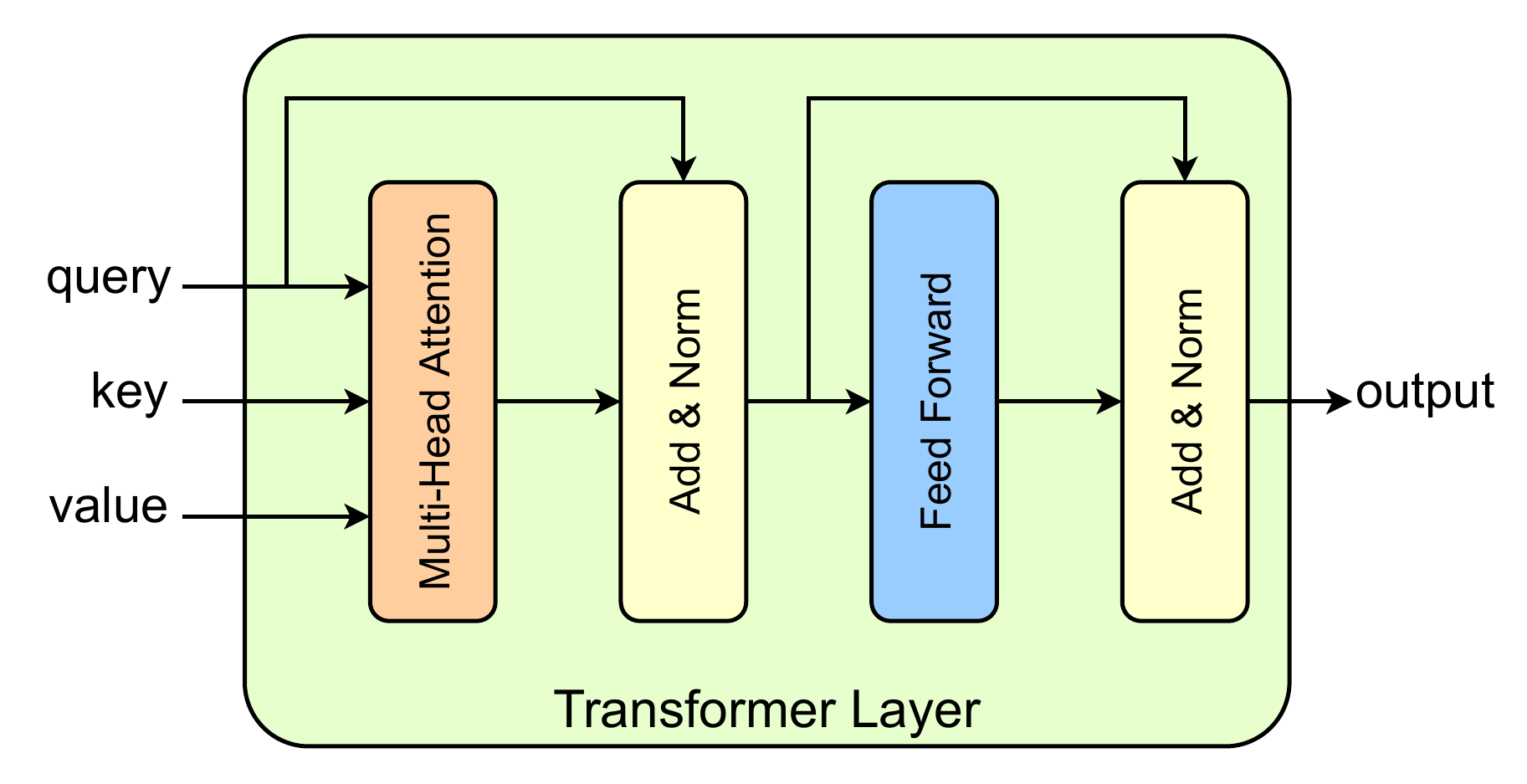}
\end{center}    
\caption{Architecture of the Transformer layer, which contain a  multi-head attention module, a feed-forward network, and two normalization layers.}
\label{fig:transformer_layer}
\end{figure}

We adopt a GPT-like multi-layer Transformer in this paper. Each transformer layer consists of a classical multi-head attention module, a feed-forward network, and normalization layers as shown in Fig. \ref{fig:transformer_layer}. 
We use the original full attention mechanism but not sparse attention in Transformers.
The feed-forward network is a two-layer perceptron, while layer normalization is used in the Transformers for normalization.

\section{Dataset details}
\label{appendix_dataset}
\begin{figure*}[http!]
\begin{center}
\includegraphics[width=0.99\linewidth]{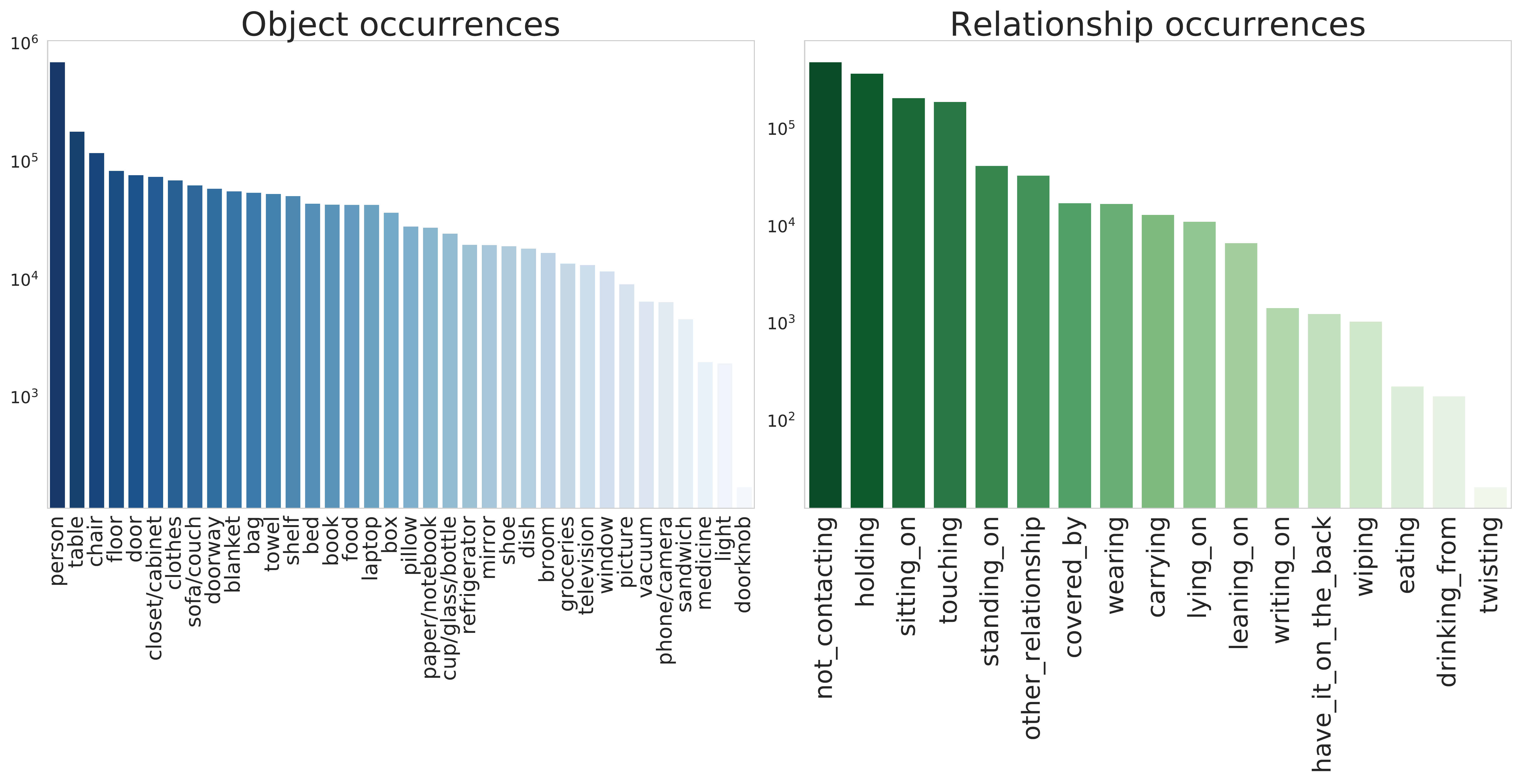}
\end{center}    
\caption{Distributions of object (left) and relationship (right) occurrences for the sub-dataset split from the Action Genome dataset.}
\label{obj_rel_occ}
\end{figure*}

We split a sub-dataset from Action Genome \cite{ji2020action}, which is built upon Charades \cite{sigurdsson2016hollywood}.
To include more complex semantic variations in the 16-frame video, we sampled 1 frame every 5 frames from the original videos of Charades and resize the sampled frames to a resolution of 128$\times$128.
We only keep the objects whose bounding boxes with short edges larger than 16 pixels.
In order to avoid overly complex scene graphs that make the representations difficult to infer, we reduce the graph fidelity by cutting out redundant nodes in the scene graph and keep a maximum of 5 object nodes.
In addition, each video contains at least 5 video scene graphs so that the video scene graph (VSG) encoder has enough information to infer the graph representations that are not given.
In the split dataset, there are 36 object categories and 17 relationship categories. 
The distribution of object and relationship occurrences are illustrated in Fig. \ref{obj_rel_occ}.

\section{Metrics details}
\label{appendix_metrics}
\paragraph{Fr\'{e}chet video distance (FVD).}
FVD \cite{unterthiner2018towards} is developed from Fr\'{e}chett Inception Distance (FID) \cite{heusel2017gans}, which is widely-used to evaluate the performance of image generation models.
FVD takes into account a distribution over entire videos in order to avoid the disadvantages of frame-level metrics.
A pre-trained Inflated 3D Convnet \cite{carreira2017quo} is used to capture video feature distributions.
The 2-Wasserstein distance between the ground truth video distribution and the synthetic video distribution is calculated as the metrics.

\paragraph{Structural similarity index measure (SSIM).}
SSIM \cite{wang2004image} is a per-frame perceptual metrics that measures the similarity between two images.
The statistical measure combines three different factors: luminance, variance and correlation.
We first split the ground truth videos and synthetic videos into single frames.
Then we calculate SSIM between the ground truth frames and synthetic frames. 
The average SSIM of all frames is taken as the final result.

\section{Technical implementation details}
\label{appendix_implementation}
\paragraph{Video scene graph representation learning framework.}
In the video scene graph encoder, both the spatial Transformer and temporal Transformer have 3 Transformer layers. 
We employ 4 attention heads for each attention module, while the dimension $d$ of the input queries, keys, and values is set to 256.
The encodings are only added to queries and keys when using the attention modules.
For the frame encoder, we adopt the CNN-based model used in \cite{xu2018attngan}, which is built upon Inception-v3 model \cite{szegedy2016rethinking}. 
The input frames are first resized to a resolution of $299\times299$, while the size of the feature maps extracted by the CNN backbone is $768\times17\times17$.
A $1\times1$ convolution layer is exploited to reduce the dimension of the feature maps to $d=256$. 
Then we use a global average pooling layer to convert the feature maps to the frame vectors.
We train the video scene graph encoder and frame encoder using ADAM optimizer \cite{kingma2014adam} with a learning rate of $1\times10^{-4}$ and a batch size of 12 images.
The training takes about 20 hours on 2 RTX 2080 TI GPUs.

\paragraph{Semantic scene graph-to-video synthesis framework.}
We adopt the VQ-VAE from \cite{le2021ccvs} and use almost the same hyperparameters. 
The encoder of the VQ-VAE converts a $128\times128$ frame into a $512\times8\times8$ feature map. 
The $8\times8$ sub-vectors of the feature map are then quantified to the discrete latent embeddings. 
The length of the latent codebook is set to 1024 to shorten the training time.
The $8\times8$ discrete latent embeddings are reconstructed to a video frame by the decoder of the VQ-VAE.
We train the VQ-VAE using ADAM optimizer \cite{kingma2014adam} with a learning rate of $2\times10^{-4}$ and a batch size of 32 videos on 8 RTX 3090 TI GPUs for about 48 hours.

The auto-regressive Transformer consists of 24 Transformer layers with a head number of 16. 
Due to the complexity of the auto-regression task, the embedding dimension of the attention module is set to 1024.
Therefore, a linear transformation is utilized to project the dimension of video scene graph representations from 256 to 1024, while 1024 latent embeddings with dimension $1024d$ are learned during the training.
We train the auto-regressive Transformer using ADAM optimizer \cite{kingma2014adam} with a learning rate of $1\times10^{-5}$ and a batch size of 64 videos on 8 RTX 3090 TI GPUs for about 48 hours. 
Furthermore, the video scene graph encoder is frozen during the training of the auto-regressive Transformer.



\begin{figure*}[http!]
\begin{center}
\includegraphics[width=0.99\linewidth]{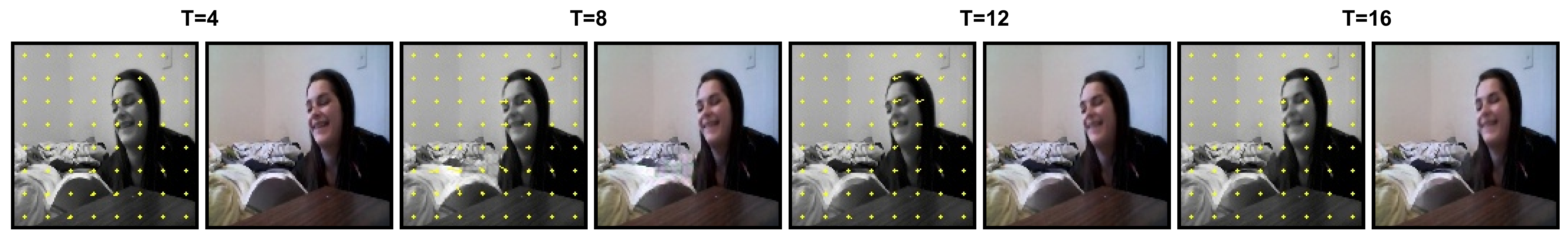}
\end{center}
\caption{Qualitative result for simple video synthesize, in which the girl keeps holding the book.
The motion in the original video is very small.
Only 4 synthesized frames and their optical flow are shown. In this case, the details such as the face are well rendered. }
\label{fig:qualitative2}
\end{figure*}

\section{Additional qualitative results and limitations}
\label{appendix_qualitative}
\paragraph{Additional qualitative results.}
The details such as the human face are not well presented in qualitative examples in the main paper. 
As discussed, the reason is that the motion in the video is quite large.
Another simple example is shown in Fig. \ref{fig:qualitative2}.
In the original video, the girl is holding and looking at the book (all the video scene graphs are the simple triplet \texttt{person-holding-book}).
Although there is some change in the position of the girl's head and book, it is not significant.
In this case, SSGVS can render better details and perform well.
The original video and some generated frames are omitted because the synthetic frames are very close to the original ones and the motion is small.
To visualize the small motion better, we also compute the optical flows for the shown synthetic frames.

\begin{figure*}[htp]
\begin{center}
\includegraphics[width=0.99\linewidth]{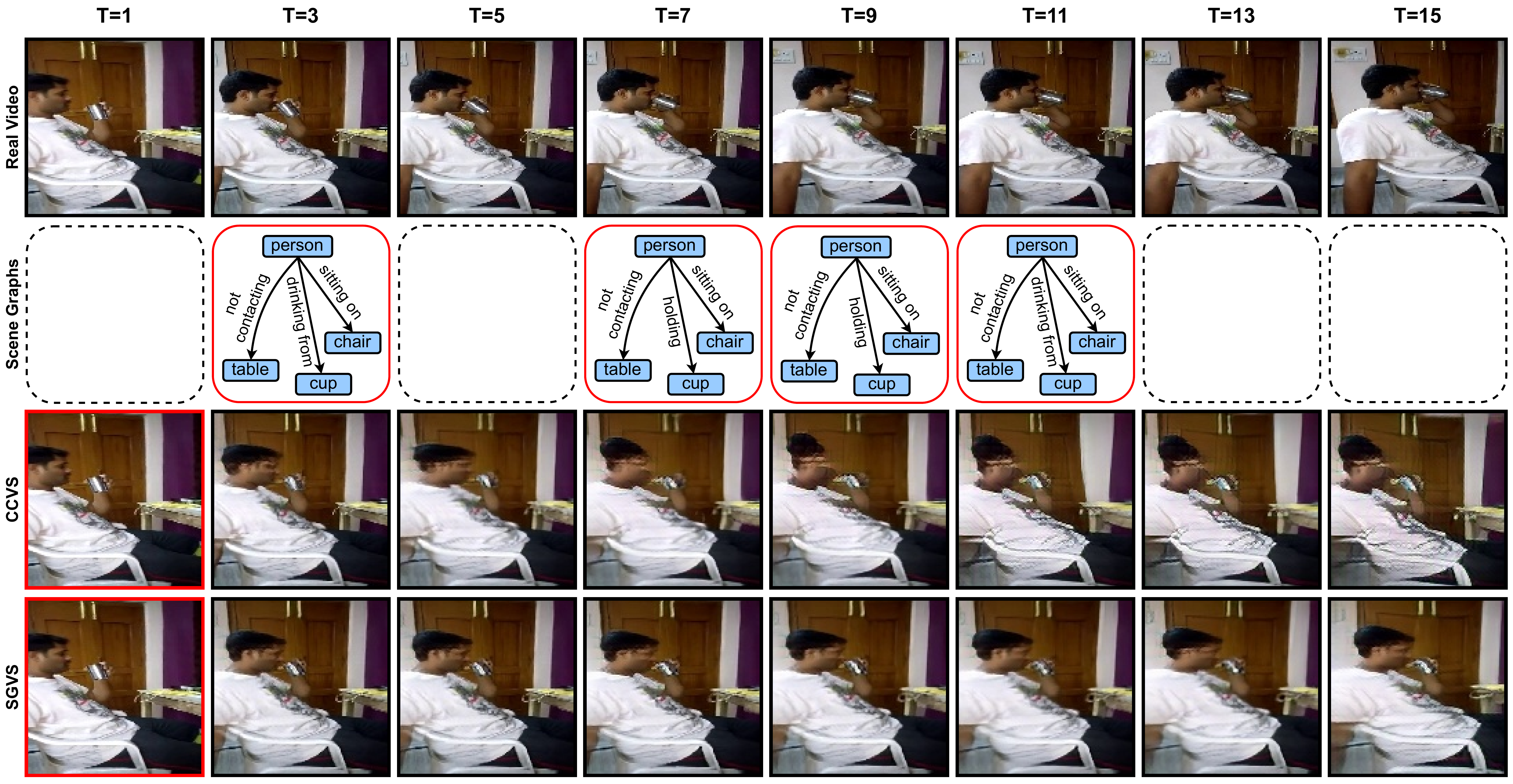}
\end{center}
\caption{Comparison between the videos synthesized by CCVS and SSGVS. The real frames are given in the first row, while the corresponding video scene graphs are shown in the second row. The video synthesized by SSGVS has higher fidelity with the help of the discrete video scene graphs.}
\label{fig:compare}
\end{figure*}

In Fig. \ref{fig:compare}, we show the video synthesized by CCVS \cite{le2021ccvs}, which only use the first frame as input, and the video synthesized by our SSGVS which use the first frame and also the video scene graphs.
With the help of the input video scene graphs, SSGVS can synthesize higher quality frames, especially those far from the starting frame.
In this example, there are no significant semantic changes in the video scene graphs. 
They control SSGVS to generate the frames that maintain the current drinking action, whereas the distortion in the frames generated by CCVS is getting worse.

\begin{figure*}[htp]
\begin{center}
\includegraphics[width=0.99\linewidth]{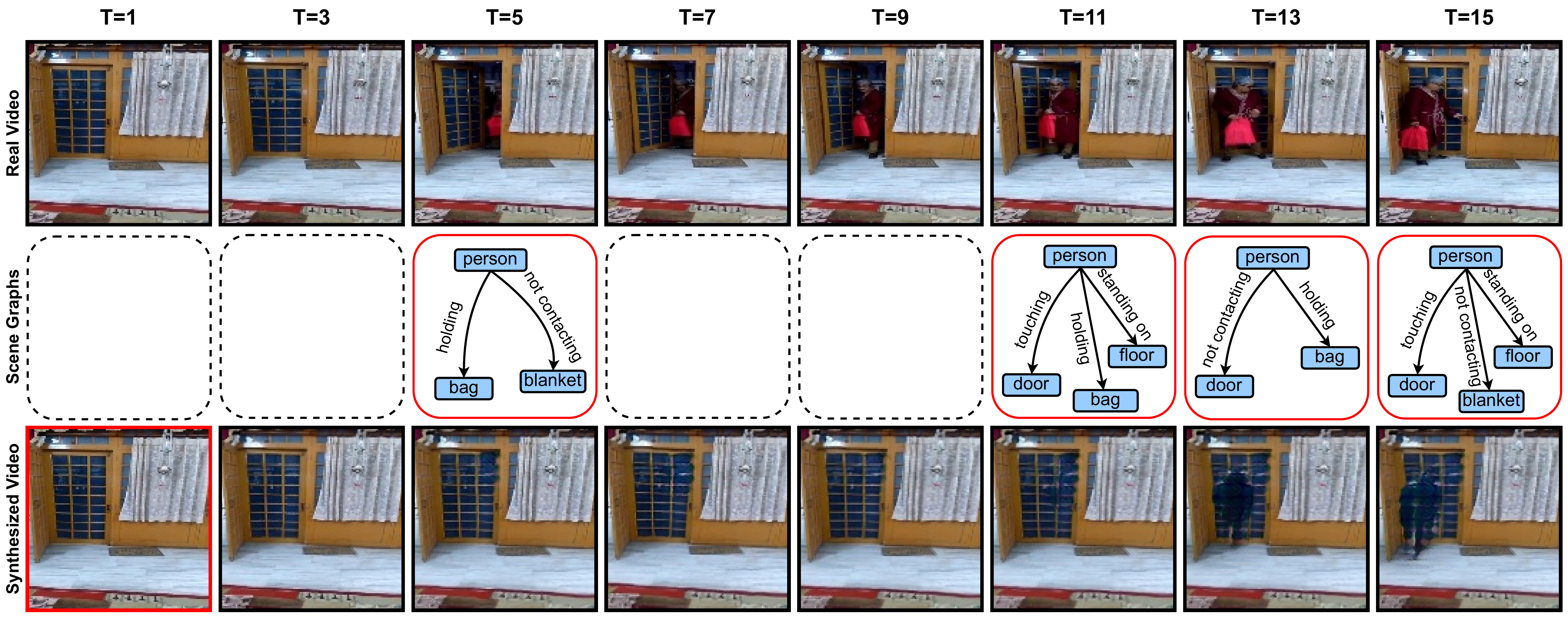}
\end{center}
  \caption{Failure to synthesize a complex video with large motion. The person and the bag cannot be synthesized since they do not appear in the first frame. In this case, only the silhouettes of a standing person are visible in the frames of $T=13$ and $T=15$.}
\label{fig:limit}
\end{figure*}

\paragraph{Limitations.}
Since the resolution of our generated video is $128\times128$, this constraint makes some small objects such as the phone and medicine cannot be presented very clearly. 
In addition, for some videos containing the large motion, the auto-regressive transformer cannot successfully predict the sequence of the latent embeddings. 
These videos usually involve a change of scene or camera pose.
An example is shown in Fig. \ref{fig:limit}.

\section{Ethics statement}
As machine learning methods are increasingly used in everyday life, it makes sense to consider the potential social impact of our work.
Our work could potentially be used for deep fake as well as other state-of-the-art generative models.
Since our model can synthesize videos with specific semantic content, this even makes deep fake more flexible.
Developing better models has the potential to be used maliciously to violate human likeness rights or create false information.
On the other hand, a good video synthesis model helps the film and video game industries, for example, by replacing live actors in dangerous scenes.
It can be also very promising in the metaverse.


{\small
\bibliographystyle{ieee_fullname}
\bibliography{egbib}
}

\end{document}